\documentclass[runningheads]{llncs}
\usepackage[T1]{fontenc}
\usepackage{graphicx}
\usepackage{color}
\usepackage{makeidx}  
\usepackage[colorlinks,linkcolor=blue]{hyperref}
\usepackage{url}
\usepackage{longtable}
\usepackage{multirow}
\usepackage{amssymb, amsmath, bm}
\usepackage{amsfonts, amssymb}
\usepackage{pifont}
\usepackage{mathtools}
\usepackage{mathrsfs}
\usepackage{booktabs}
\usepackage{cite}
\usepackage{dsfont}
\usepackage{makecell}
\usepackage[noend]{algpseudocode}
\usepackage{algorithmicx,algorithm}
\usepackage[misc]{ifsym} 
\usepackage{bbding}
\usepackage{ulem}
\newcommand{\M}{CSG-3DCT}

\begin{document}

\title{Inflated 3D Convolution-Transformer for Weakly-supervised Carotid Stenosis Grading with Ultrasound Videos}

\titlerunning{Inflated 3D Convolution-Transformer for Weakly-supervised CSG}

\author{Xinrui Zhou\inst{1,2,3}\thanks{Xinrui Zhou and Yuhao Huang contribute equally to this work.}, Yuhao Huang\inst{1,2,3\star} \and Wufeng Xue\inst{1,2,3} \and Xin Yang\inst{1,2,3} \and \\Yuxin Zou\inst{1,2,3} \and Qilong Ying\inst{1,2,3} \and Yuanji Zhang\inst{1,2,3,4} \and Jia Liu\inst{5} \and Jie Ren\inst{5} \and Dong Ni\inst{1,2,3}\textsuperscript{(\Letter)}} 

\institute{
\textsuperscript{$1$}National-Regional Key Technology Engineering Laboratory for Medical Ultrasound, School of Biomedical Engineering, Health Science Center, Shenzhen University, China\\
\email{nidong@szu.edu.cn} \\
\textsuperscript{$2$}Medical Ultrasound Image Computing (MUSIC) Lab, Shenzhen University, China\\
\textsuperscript{$3$}Marshall Laboratory of Biomedical Engineering, Shenzhen University, China\\
\textsuperscript{$4$}Shenzhen RayShape Medical Technology Co., Ltd, China\\
\textsuperscript{$5$}The Third Affiliated Hospital, Sun Yat-sen University, China\\}

\authorrunning{Zhou et al.}
\maketitle              

\begin{abstract}
Localization of the narrowest position of the vessel and corresponding vessel and remnant vessel delineation in carotid ultrasound (US) are essential for carotid stenosis grading (CSG) in clinical practice.
However, the pipeline is time-consuming and tough due to the ambiguous boundaries of plaque and temporal variation. To automatize this procedure, a large number of manual delineations are usually required, which is not only laborious but also not reliable given the annotation difficulty.
In this study, we present the first video classification framework for automatic CSG.
Our contribution is three-fold. 
First, to avoid the requirement of laborious and unreliable annotation, we propose a novel and effective video classification network for weakly-supervised CSG.  
Second, to ease the model training, we adopt an inflation strategy for the network, where pre-trained 2D convolution weights can be adapted into the 3D counterpart in our network for an effective warm start. 
Third, to enhance the feature discrimination of the video, we propose a novel attention-guided multi-dimension fusion (AMDF) transformer encoder to model and integrate global dependencies within and across spatial and temporal dimensions, where two lightweight cross-dimensional attention mechanisms are designed.
Our approach is extensively validated on a large clinically collected carotid US video dataset, demonstrating state-of-the-art performance compared with strong competitors.
\end{abstract}

\section{Introduction}
\begin{figure*}[!t]
	\centering
	\includegraphics[width=0.7\linewidth]{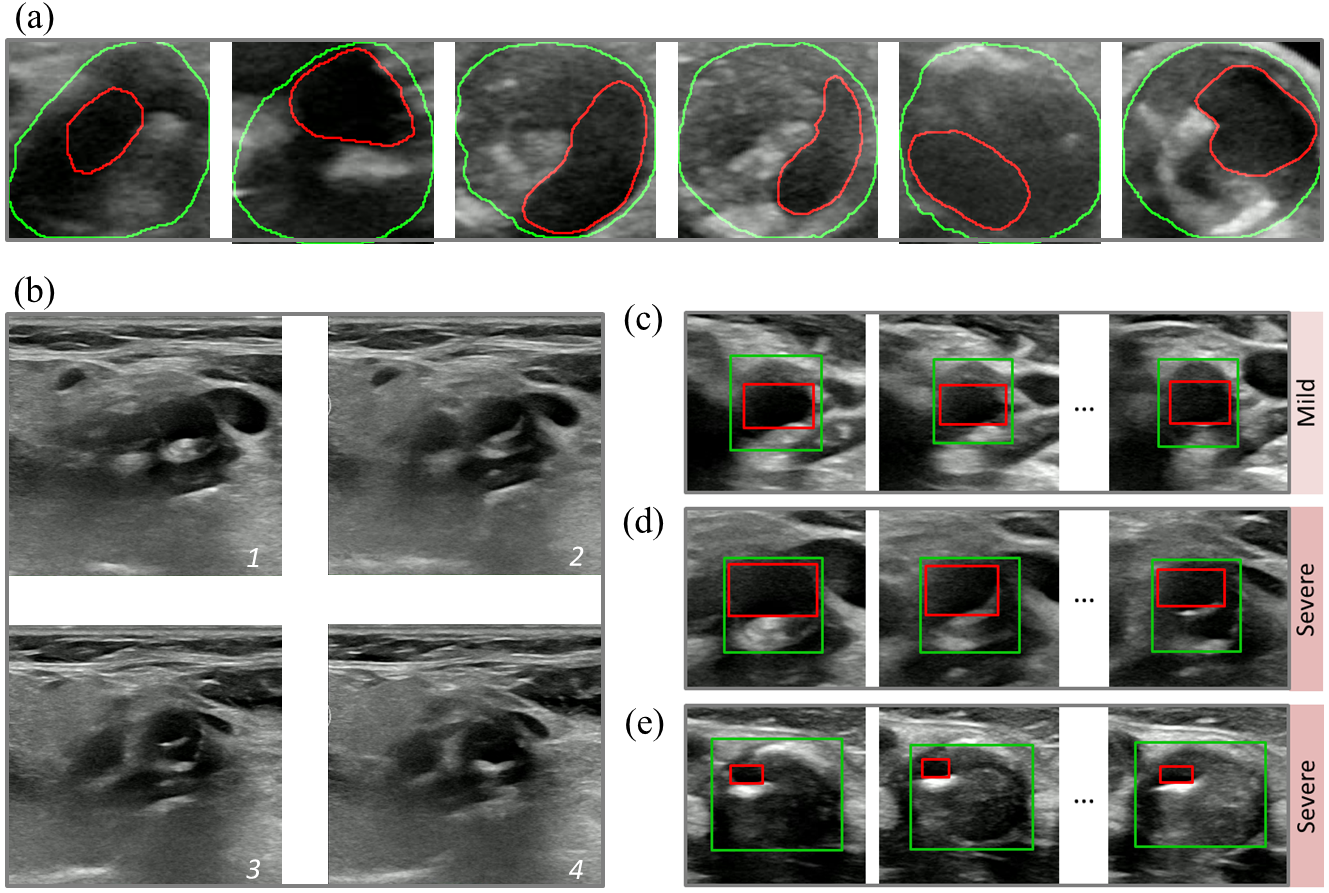}
	\caption{(a) Carotid plaque images in US with annotated vessel (green contour) and remnant vessel (red contour). (b) An original video with vessels and plaques dynamically displayed. (c-e): Cropped plaque clips with annotated vessel (green box), remnant vessel (red box), and corresponding label.}
	\label{fig:intro}
\end{figure*}

Carotid stenosis grading (CSG) represents the severity of carotid atherosclerosis, which is highly related to stroke risk~\cite{howard2021risk}.
In clinical practice, sonographers need to first visually locate the frame with the largest degree of vascular stenosis (i.e., minimal area of remnant vessels) in a dynamic plaque video clip based on B-mode ultrasound (US), then manually delineate the contours of both vessels and remnant vessels on it to perform CSG.
However, the two-stage pipeline is time-consuming and the diagnostic results heavily rely on operator experience and expertise due to ambiguous plaque boundaries and temporal variation (see Fig.~\ref{fig:intro}).
Fully-supervised segmentation models can automatize this procedure, but require numerous pixel-level masks laboriously annotated by sonographers and face the risk of training failure due to unreliable annotation. 
Hence, tackling this task via weak supervision, i.e., video classification, is desired to avoid the requirement of tedious and unreliable annotation.

Achieving accurate automatic CSG with US videos is challenging. First, the plaque clips often have extremely high intra-class variation due to changeable plaque echo intensity, shapes, sizes, and positions (Fig.~\ref{fig:intro}(d-e)). The second challenge lies in the inter-class similarity of important measurement indicators (i.e., diameter and area stenosis rate) for CSG among cases with borderlines of mild and severe, which makes designing automatic algorithms difficult (Fig.~\ref{fig:intro}(c-d)).

A typical approach for this video classification task is CNN-LSTM~\cite{donahue2015long}. Whereas, such \textit{2D + 1D} paradigm lacks interaction with temporal semantics of input frames in the early stage. Instead, a more efficient way is to build 3D networks that handle spatial and temporal (ST) information simultaneously~\cite{yang2020temporal}.

There are several types of 3D networks that have been widely used in visual tasks: (1) Pure 3D convolution neural networks (3D CNNs) refer to capturing local ST features using convolution operations~\cite{feichtenhofer2019slowfast, yang2020temporal, carreira2017quo}.
However, most current 3D CNNs suffer from the lack of good initialization and capacity for extracting global representations~\cite{peng2021conformer}. (2) Pure 3D transformer networks (3D Trans) aim to exploit global ST features by applying self-attention mechanisms~\cite{bertasius2021space, fan2021multiscale, liu2022video, yan2022multiview}.
However, their ability in extracting local ST information is weaker than 3D CNNs. Moreover, such designs have not deeply explored lightweight cross-dimensional attention mechanisms to gain refined fused features for classification.

Recently, Wang et al.~\cite{wang2018non} first introduced the self-attention mechanism in 3D CNN for video classification. ~\cite{peng2021conformer, xu2021vitae} then proposed Convolution-Transformer hybrid networks for image classification. Li et al.~\cite{li2022uniformer} further extended such hybrid design to 3D  for video recognition by seamlessly integrating 3D convolution and self-attention. Thanks to both operations, such networks can fully exploit and integrate local and global features, and thus achieve state-of-the-art results.
However, current limited 3D hybrid frameworks are designed in cascade, which may lead to semantic misalignment between CNN- and Transformer-style features and thus degrade accuracy of video classification.

In this study, we present the first video classification framework based on 3D Convolution-Transformer design for CSG (named \M).
Our contribution is three-fold. 
First, we propose a novel and effective video classification network for weakly-supervised CSG, which can avoid the need of laborious and unreliable mask annotation.
Second, we adopt an inflation strategy to ease the model training, where pre-trained 2D convolution weights can be adapted into the 3D counterpart. In this case, our network can implicitly gain the pre-trained weights of existing large models to achieve an effective warm start.
Third, we propose a novel play-and-plug attention-guided multi-dimension fusion (AMDF) transformer encoder to integrate global dependencies within and across ST dimensions. Two lightweight cross-dimensional attention mechanisms are devised in AMDF to model ST interactions, which merely use class (CLS) token~\cite{dosovitskiy2021an} as Query.
Extensive experiments show that \M~achieve state-of-the-art performance in CSG task.

\section{Methodology}
\begin{figure*}[!t]
    \centering
    \includegraphics[width=1.0\linewidth]{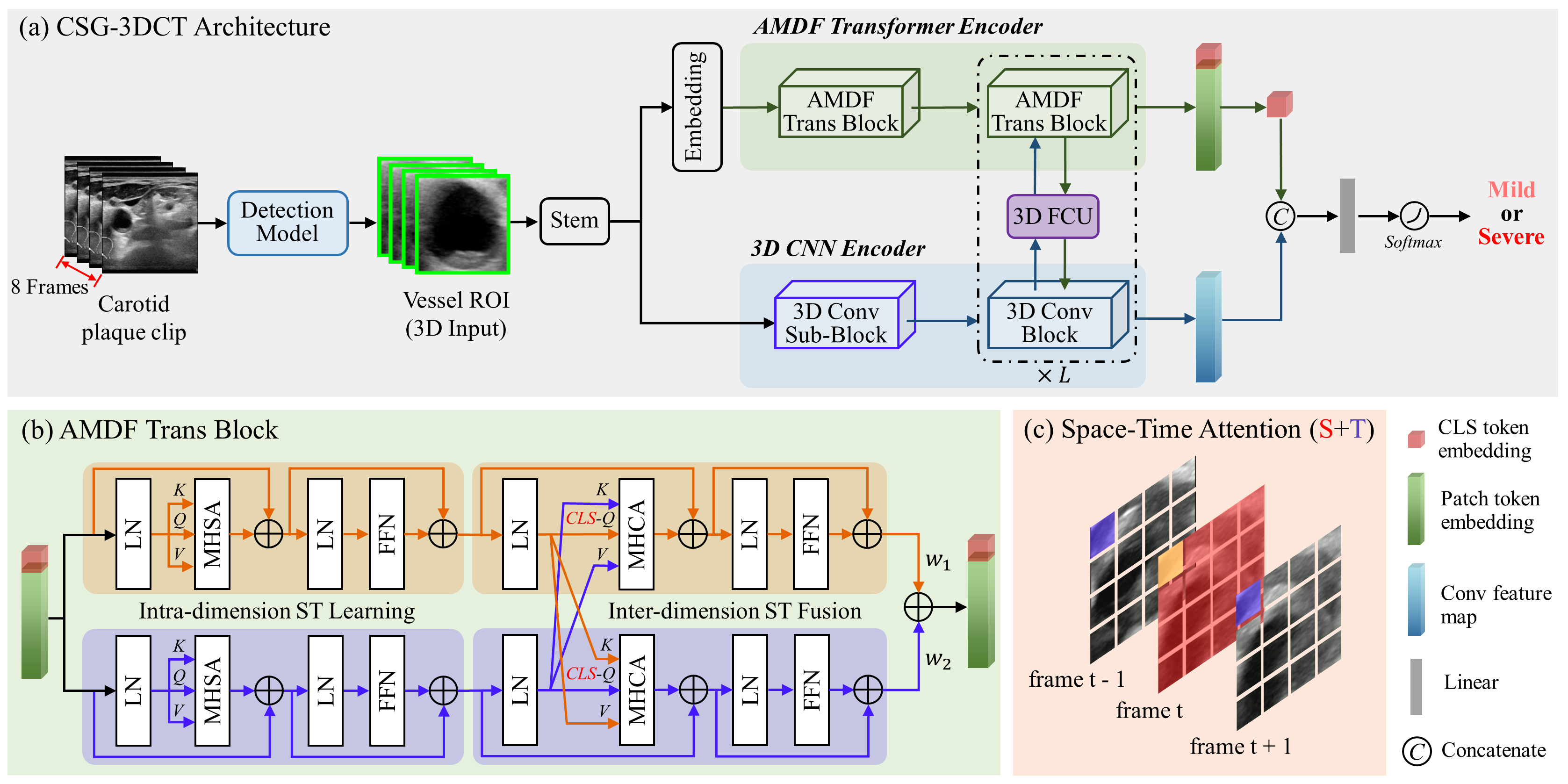}
    \caption{(a) Overview of \M. It contains \textit{L+1} and \textit{L} repeated AMDF Trans and 3D Conv Blocks, respectively. A 3D Conv Block consists of two sub-blocks. (b) Pipeline of the AMDF Trans Block. (c) Visualization of the space-time attention used in the intra-dimension ST learning module. Yellow, blue and red patches indicate query, and attention separately adopted along temporal and spatial dimensions, respectively.} 
    \label{fig:framework}
\end{figure*}

Fig.~\ref{fig:framework}(a) shows the pipeline of our proposed framework.
Note that the proposed \M~is inflated from the 2D architecture. 
Thus, it can implicitly gain the pre-trained weights of current large model for effective initialization.
In \M, given a video clip, vessel regions are first detected by the pre-trained detection model~\cite{liu2022deep} for reducing redundant background information.
Then, the cropped regions are concatenated to form a volumetric vessel and input to the 3D CNN and Transformer encoders.
Specifically, to better model the global knowledge, ST features are decoupled and fused by the proposed AMDF transformer encoder.
CNN- and Transformer-style features are integrated by the 3D feature coupling unit (3D FCU)~\cite{peng2021conformer} orderly.
Finally, by combining the CNN features and the CLS token, the model will output the label prediction.

\subsubsection{3D Mix-architecture for Video Classification.}
CNN and Transformer have been validated that they specialize in extracting local and global features, respectively.
Besides, compared to the traditional 2D video classifiers, 3D systems have shown the potential to improve classification accuracy due to their powerful capacity of encoding multi-dimensional information.
Thus, in \M, we propose to leverage the advantages of both CNN and Transformer and extend the whole framework to a 3D version. 

The meta-architecture of our proposed \M~follows the 2D Convolution-Transformer (Conformer) model~\cite{peng2021conformer}.
It mainly has 5 stages (termed \textit{c1-c5}).
Extending it to 3D represents that both CNN and Transformer should be modified to adapt the 3D input.
In specific, we tend to inflate the 2D \textit{k$\times$k} convolution kernels to 3D ones with the size of \textit{$t\times{k^2}$} by adding a temporal dimension, which is similar to~\cite{carreira2017quo}.
Such kernels can be implicitly pre-trained on ImageNet through bootstrapping operation\cite{carreira2017quo}.
While translating the 2D transformer only requires adjusting the token number according to the input dimension.

\subsubsection{Inflation Strategy for 3D CNN Encoder.}
We devise an inflation strategy for the 3D CNN encoder to relieve the model training and enhance the representation ability. 
For achieving 2D-to-3D inflation, a feasible scheme is to expand all the 2D convolution kernels at temporal dimension with \textit{t>1}~\cite{carreira2017quo}.
However, multi-temporal (\textit{t>1}) 3D convolutions are computationally complex and hard to train. 
Thus, we only select part of the convolution kernels for inflating their temporal dimension larger than 1, while others restrict the temporal dimension to 1.
By adapting pre-trained 2D convolution weights into the 3D counterpart, our network can achieve good initialization from existing large model.
Moreover, we notice that performing convolutions at a temporal level in early layers may degrade accuracy due to the over-neglect of spatial learning~\cite{feichtenhofer2019slowfast}.
Therefore, instead of taking a whole-stage temporal convolution, we only perform it on 3D Conv blocks of the last three stages (i.e., \textit{c3-c5}).
We highlight that our temporal convolutions are length-invariant, which indicates that we will not down-sample at the temporal dimension. 
It can benefit the maintenance of both video fidelity and time-series knowledge, especially for short videos.
More details refer to supplementary material (Table~\ref{tab:supp}).

\subsubsection{Transformer Encoder with Play-and-plug AMDF design.} 
Simply translating the 2D transformer encoder into the 3D standard version mainly has two limitations:
(1) It blindly compares the similarity of all ST tokens by self-attention, which tends to inaccurate predictions. Moreover, such video-based computation handles \textit{t$\times$} tokens simultaneously compared to image-based methods, leading to much computational cost.
(2) It also has no ability to decide which information is more important during different learning stages.
Thus, we propose to enhance the decoupled ST features and their interactions using different attention manners.
The proposed encoder can improve computational efficiency, and can be flexibly integrated into 2D or 3D transformer-based networks.

Before the transformer encoder, we first decompose the feature maps \textit{X} produced by the stem module into $t\times{n^2}$ embeddings without overlap.
A CLS token \textit{X$_{cls}$}$\in\mathbb{R}^d$ is then added in the start position of \textit{X} to obtain merged embeddings \textit{Z}$\in\mathbb{R}^{d\times(t\times{n^2}+1)}$.
$n^2$ and $d$ denote the number of spatial patch tokens and hidden dimensions, respectively.
Then, the multiple AMDF Trans blocks in the transformer encoder drive \textit{Z} to produce multi-dimensional enhanced representations. 
Specifically, the AMDF block has the following main components.

\textbf{1) Intra-dimension ST Learning Module.}
Different from the cascade structure in~\cite{bertasius2021space}, \M~constructs two parallel branches to learn global ST features, respectively.
As shown in Fig.~\ref{fig:framework}(b), the proposed module is following ViT~\cite{dosovitskiy2021an}, which consists of a multi-head self-attention (MHSA) module and a feed-forward network (FFN). 
Query-Key-Value (QKV) projection after LayerNorms~\cite{ba2016layer} is conducted before each MHSA module. 
Besides, the residual connections are performed in MHSA module and FFN. 
Taking token embeddings as input, the two branches can extract the ST features well by parallel spatial and temporal attention (see Fig.~\ref{fig:framework}(c) for visualization of computation process).

\begin{figure*}[!t]
    \centering
    \includegraphics[width=0.7\linewidth]{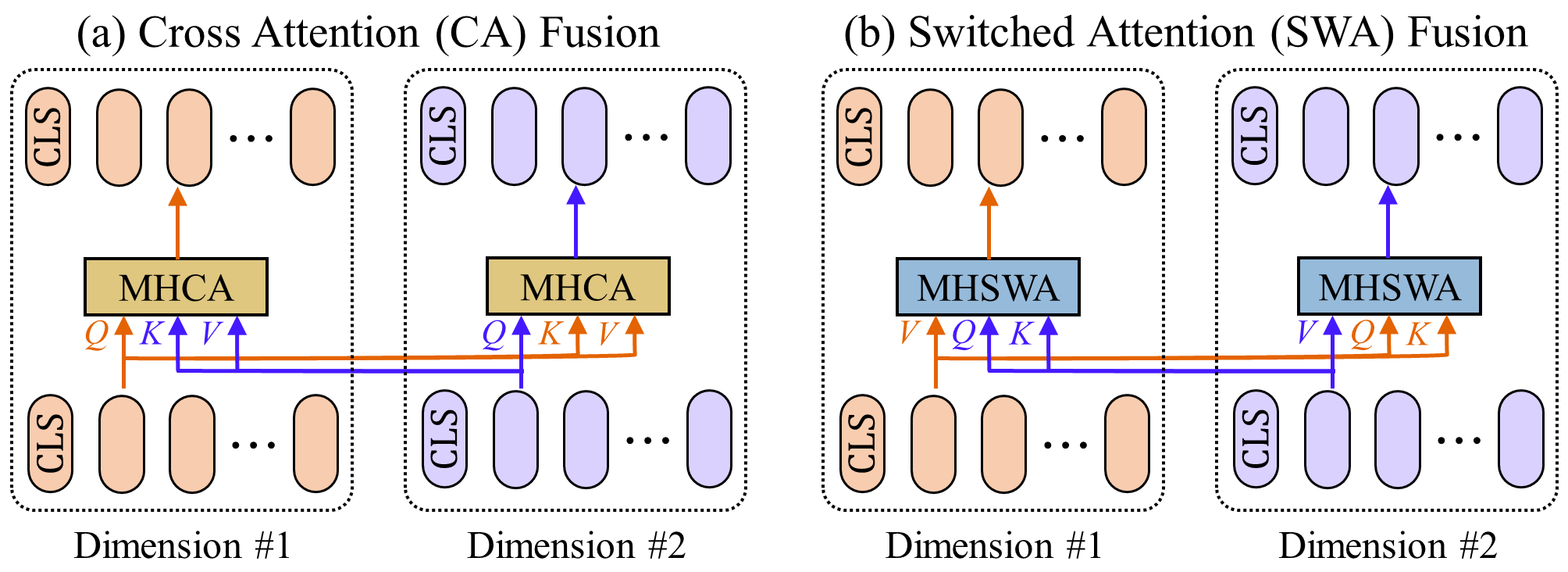}
    \caption{An illustration of our proposed multi-dimension fusion methods.} 
    \label{fig:fusion_illustration}
\end{figure*}

\textbf{2) Inter-dimension ST Fusion Module.}
To boost interactions between S and T dimensions, we build the inter-dimension fusion module after the intra-dimension learning module.
The only difference between the two types of modules is the calculation mode of attention.
As shown in Fig.~\ref{fig:fusion_illustration}, we consider the following two methods to interact the ST features: (i) \textit{Switched Attention (SWA) Fusion} and (ii) \textit{Cross Attention (CA) Fusion.}
Here, we define one branch as the target dimension and the other branch as the complementary dimension.
For example, when the temporal features are flowing to the spatial features, the spatial branch is the target and the temporal branch is the complementary one.
SWA is an intuitive way for information interaction.
It uses the attention weights (i.e., generated by $\mathbf{Q}$ and $\mathbf{K}$) as the bridge to directly swap the information.
For computing the target features in CA, $\mathbf{K}$ and $\mathbf{V}$ are from the complementary one, and $\mathbf{Q}$ is from its own.
Intuition behind CA is that the target branch can \textit{query} the useful information from the given $\mathbf{K}$ and $\mathbf{V}$~\cite{curto2021dyadformer}.
Thus, the \textit{querying} process in CA can better encourage the knowledge flowing. 

To improve computing efficiency in CA, Chen et al.~\cite{chen2021crossvit} proposed to adopt the CLS token of the target branch to compute the CLS-$\mathbf{Q}$ to replace the common $\mathbf{Q}$ from token embeddings. Then, they transferred the target CLS token to the complementary branch to obtain the $\mathbf{K}$ and $\mathbf{V}$ and perform CA. However, such a design may lead to overfitting due to the query-queried feature dependency. Motivated by~\cite{chen2021crossvit}, we introduce a simple yet efficient attention strategy in inter-dimension ST fusion module. Specifically, the target dimension adopts its CLS token as a \textit{query} to mine rich information, and this CLS token will not be inserted into the complementary dimension. Besides, using one token only can reduce the computation time quadratically compared to all tokens attention. 

\textbf{3) Learnable Mechanism for Adaptive Updating.}
Multi-dimensional features commonly have distinct degrees of contribution for prediction. For example, supposing the size of carotid plaque does not vary significantly in a dynamic segment, the spatial information may play a dominant role in making the final diagnosis. 
Thus, we introduce a learnable parameter to make the network adaptively adjust the weights of different branches and learn the more important features (see Fig.~\ref{fig:framework}(b)). 
We highlight that this idea is easy to implement and general to be equipped with any existing feature-fusion modules.

\section{Experimental Results}
\begin{table}[!t]
    \centering
    \caption{Quantitative results of methods. "MTV(B/2+S/8)" means to use the larger "B" model to encode shorter temporal information (2 frames), and the smaller "S" model to encode longer temporal information (8 frames)~\cite{yan2022multiview}. $\dagger$ denotes random initialization. $\ast$ indicates removing the learnable mechanism from AMDF encoder.}
    \setlength{\tabcolsep}{3mm}
    \begin{tabular}{c|c|c|c|c}
    \toprule
    Methods & Accuracy & F1-score & Precision & Recall \\
    \hline
    
    I3D~\cite{carreira2017quo} & 78.8$\%$ & 78.8$\%$ & 81.0$\%$ & 80.8$\%$ \\ 
    SlowFast~\cite{feichtenhofer2019slowfast} & 70.3$\%$ & 70.2$\%$ & 70.3$\%$ & 70.8$\%$ \\ 
    TPN~\cite{yang2020temporal} & 78.0$\%$ & 77.8$\%$ & 77.8$\%$ & 78.5$\%$ \\ 
    \hline
    
    TimeSformer~\cite{bertasius2021space} & 77.1$\%$ & 76.2$\%$ & 76.8$\%$ & 75.9$\%$ \\ 
    Vivit~\cite{arnab2021vivit} & 70.3$\%$ & 70.2$\%$ & 70.3$\%$ & 70.8$\%$ \\ 
    MTV (B/2+S/8)~\cite{yan2022multiview} & 72.0$\%$ & 72.0$\%$ & 73.0$\%$ & 73.4$\%$ \\ 
    \hline
    
    NL I3D~\cite{wang2018non} & 78.8$\%$ & 78.8$\%$ & 81.0$\%$ & 80.8$\%$ \\ 
    UniFormer~\cite{li2022uniformer} & 75.4$\%$ & 75.4$\%$ & 78.1$\%$ & 77.6$\%$ \\
    \hline

    CSG-3DCT-Base & 80.5$\%$ & 80.2$\%$ & 80.0$\%$ & 80.4$\%$ \\
    CSG-3DCT-Base$^\dagger$ & 76.3$\%$ & 75.4$\%$ & 75.8$\%$ & 75.2$\%$ \\
    CSG-3DCT-Base-16 & 79.8$\%$ & 78.0$\%$ & 79.3$\%$ & 77.4$\%$ \\
    CSG-3DCT-SWA$^\ast$ & 82.2$\%$ & 81.4$\%$ & 82.5$\%$ & 80.9$\%$ \\
    CSG-3DCT-CA$^\ast$ & 82.2$\%$ & 82.1$\%$ & 82.2$\%$ & \textcolor{blue}{83.0$\%$}\\
    CSG-3DCT & \textcolor{blue}{83.1$\%$} & \textcolor{blue}{82.5$\%$} & \textcolor{blue}{82.8$\%$} & 82.4$\%$ \\
    \bottomrule
    \end{tabular}
\label{tab:results}
\end{table}

\subsubsection{Dataset and Implementations.}
We validated the \M~on a large in-house carotid transverse US video dataset. Approved by the local IRB, a total of 200 videos (63225 images with size 560$\times$560 and 380$\times$380) were collected from 169 patients with carotid plaque. In clinic, sonographers often focus on a relatively narrow short plaque video clip instead of the long video. Thus, we remade the dataset by using the key plaque video clips instead of original long videos. 
Specifically, sonographers with 7-year experience manually annotated 8/16 frames for a plaque clip and labeled the corresponding stenosis grading (mild/severe) using the Pair annotation software package~\cite{liang2022sketch}. The final dataset was split randomly into 318, 23, and 118 plaque clips with 8 frames or into 278, 23, and 109 ones with 16 frames for training, validation, and independent testing set at the patient level with no overlap.

In this study, we implemented \M~in \textit{Pytorch}, using an NVIDIA A40 GPU. Unless specified, we trained our model using 8-frame input plaque clips.  All frames were resized to 256 $\times$ 256.  The learnable weights of QKV projection and LayerNorm weights in spatial dimension branch of intra-dimension ST learning module were initialized with those from transformer branch in Conformer~\cite{peng2021conformer}, while other parameters in AMDF transformer encoder performed random initialization. We trained \M~using Adam optimizer with the learning rate (\textit{lr}) of 1e-4 and weight decay of 1e-4 for 100 epochs. Batch size was set as 4. Inspired by~\cite{wang2018non}, \M~with 16-frame inputs was initialized with 8-frame model and fine-tuned using an initial \textit{lr} of 0.0025 for 40 epochs. 

\subsubsection{Quantitative and Qualitative Analysis.}
We conducted extensive experiments to evaluate \M. 
Accuracy, F1-score, precision and recall were evaluation metrics. Table~\ref{tab:results} compares \M~with other 8 strong competitors, including 3D CNNs, 3D Trans, and 3D Mix-architecture. Note that "-Base" is directly inflated from Conformer~\cite{peng2021conformer}. Among all the competitors, -Base achieves the best results on accuracy and f1-score. It can also be observed that our proposed \M~achieves state-of-the-art results (at least \textbf{4.3$\%$} improvement in accuracy). 

\begin{figure*}[!t]
    \centering
    \includegraphics[width=0.85\linewidth]{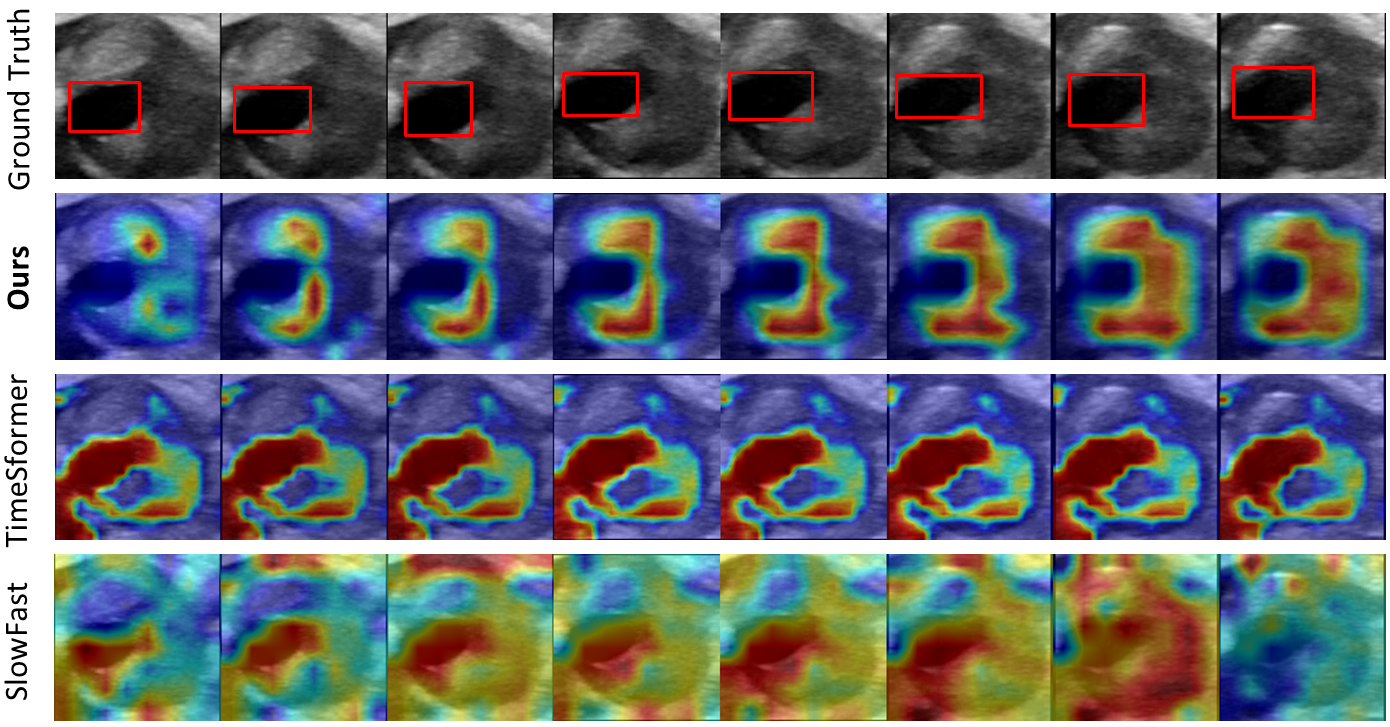}
    \caption{Attention maps of one carotid severe stenosis testing case (shown in cropped volumetric vessel). Red box denotes remnant vessel annotated by sonographers.} 
    \label{fig:visualization}
\end{figure*}

Fig.~\ref{fig:visualization} visualizes feature maps of different typical networks using Grad-CAM~\cite{selvaraju2017grad}. We use models without temporal downsampling (i.e., TimeSformer~\cite{bertasius2021space} and the "fast" branch of SlowFast~\cite{feichtenhofer2019slowfast}) to observe attention changes along temporal dimension. Both models ignore capturing equally important local and global ST features simultaneously, resulting in imprecise and coarse attention to the key object, i.e., the plaque area. Compared to both cases, \M~can progressively learn the ST contexts in an interactive fashion. As a result, the attention area is more accurate and complete, indicating the stronger discriminative ability of the learned features by \M, which proves the efficacy of our framework. 

\subsubsection{Ablation Study.}
We performed ablation experiments in the last 6 rows of Table~\ref{tab:results}. "-SWA$^\ast$" uses SWA in AMDF transformer encoder, while "-CA$^\ast$" uses CA instead. "-Base-16" denotes our "-Base" model with 16-frame inputs.

\textbf{1) Effects of Different Key Components of Our Model Design.}
We compared \M~with three variants (i.e., -Base, -SWA$^\ast$, and -CA$^\ast$) to analyze the effects of different key components. Compared with -Base, each of our proposed modules and their combination can help improve the accuracy. We adopt CA in our final model for its good performance. 

\textbf{2) Effects of Plaque Clip Length.}
We only investigated the effects of our model on 8-frame and 16-frame input clips due to limited GPU memory. We can find in Table~\ref{tab:results} that longer input clips slightly degrade the performance. This is reasonable since the frame-extracting method has been applied in the original videos, causing the covered range of plaque from a longer plaque clip is relatively wider, which is not beneficial to stenosis grading.

\textbf{3) Effectiveness of Initialization with ImageNet.}
We evaluated the value of training models starting from ImageNet-pretrained weights compared with scratch. It can be seen in Table~\ref{tab:results} that model with pretraining significantly boosts \textbf{+4.2$\%$} Acc., demonstrating the efficacy of good initialization.

\section{Conclusion}
We propose a novel and effective video classification network for automatic weakly-supervised CSG. 
To the best of our knowledge, this is the first work to tackle this task. 
By adopting an inflation strategy, our network can achieve effective warm start and make more accurate predictions. 
Moreover, we develop a novel AMDF Transformer encoder to enhance the feature discrimination of the video with reduced computational complexity. 
Experiments on our large in-house dataset demonstrate the superiority of our method. 
In the future, we will explore to validate the generalization capability of \M~on more large datasets and extend two-grade classification to four-grade of carotid stenosis.

\subsubsection{Acknowledgements.}
This work was supported by the grant from National Natural Science Foundation of China (Nos. 62171290, 62101343), Shenzhen-Hong Kong Joint Research Program (No. SGDX20201103095613036), Shenzhen Science and Technology Innovations Committee (No. 20200812143441001), Shenzhen College Stable Support Plan (Nos. 20220810145705001, 20200812162245001), and National Natural Science Foundation of China (No 81971632).

\bibliographystyle{splncs04}
\bibliography{reference}

\begin{thebibliography}{10}
\providecommand{\url}[1]{\texttt{#1}}
\providecommand{\urlprefix}{URL }
\providecommand{\doi}[1]{https://doi.org/#1}

\bibitem{arnab2021vivit}
Arnab, A., Dehghani, M., Heigold, G., Sun, C., Lu{\v{c}}i{\'c}, M., Schmid, C.:
  Vivit: A video vision transformer. In: Proceedings of the IEEE/CVF
  international conference on computer vision. pp. 6836--6846 (2021)

\bibitem{ba2016layer}
Ba, J.L., Kiros, J.R., Hinton, G.E.: Layer normalization. arXiv preprint
  arXiv:1607.06450  (2016)

\bibitem{bertasius2021space}
Bertasius, G., Wang, H., Torresani, L.: Is space-time attention all you need
  for video understanding? In: ICML. vol.~2, p.~4 (2021)

\bibitem{carreira2017quo}
Carreira, J., Zisserman, A.: Quo vadis, action recognition? a new model and the
  kinetics dataset. In: proceedings of the IEEE Conference on Computer Vision
  and Pattern Recognition. pp. 6299--6308 (2017)

\bibitem{chen2021crossvit}
Chen, C.F.R., Fan, Q., Panda, R.: Crossvit: Cross-attention multi-scale vision
  transformer for image classification. In: Proceedings of the IEEE/CVF
  international conference on computer vision. pp. 357--366 (2021)

\bibitem{curto2021dyadformer}
Curto, D., Clap{\'e}s, A., Selva, J., Smeureanu, S., Junior, J., Jacques, C.,
  Gallardo-Pujol, D., Guilera, G., Leiva, D., Moeslund, T.B., et~al.:
  Dyadformer: A multi-modal transformer for long-range modeling of dyadic
  interactions. In: Proceedings of the IEEE/CVF international conference on
  computer vision. pp. 2177--2188 (2021)

\bibitem{donahue2015long}
Donahue, J., Anne~Hendricks, L., Guadarrama, S., Rohrbach, M., Venugopalan, S.,
  Saenko, K., Darrell, T.: Long-term recurrent convolutional networks for
  visual recognition and description. In: Proceedings of the IEEE conference on
  computer vision and pattern recognition. pp. 2625--2634 (2015)

\bibitem{dosovitskiy2021an}
Dosovitskiy, A., Beyer, L., Kolesnikov, A., Weissenborn, D., Zhai, X.,
  Unterthiner, T., Dehghani, M., Minderer, M., Heigold, G., Gelly, S.,
  Uszkoreit, J., Houlsby, N.: An image is worth 16x16 words: Transformers for
  image recognition at scale. In: International Conference on Learning
  Representations (2021)

\bibitem{fan2021multiscale}
Fan, H., Xiong, B., Mangalam, K., Li, Y., Yan, Z., Malik, J., Feichtenhofer,
  C.: Multiscale vision transformers. In: Proceedings of the IEEE/CVF
  International Conference on Computer Vision. pp. 6824--6835 (2021)

\bibitem{feichtenhofer2019slowfast}
Feichtenhofer, C., Fan, H., Malik, J., He, K.: Slowfast networks for video
  recognition. In: Proceedings of the IEEE/CVF international conference on
  computer vision. pp. 6202--6211 (2019)

\bibitem{howard2021risk}
Howard, D.P., Gaziano, L., Rothwell, P.M.: Risk of stroke in relation to degree
  of asymptomatic carotid stenosis: a population-based cohort study, systematic
  review, and meta-analysis. The Lancet Neurology  \textbf{20}(3),  193--202
  (2021)

\bibitem{li2022uniformer}
Li, K., Wang, Y., Peng, G., Song, G., Liu, Y., Li, H., Qiao, Y.: Uniformer:
  Unified transformer for efficient spatial-temporal representation learning.
  In: International Conference on Learning Representations (2022)

\bibitem{liang2022sketch}
Liang, J., Yang, X., Huang, Y., Li, H., He, S., Hu, X., Chen, Z., Xue, W.,
  Cheng, J., Ni, D.: Sketch guided and progressive growing gan for realistic
  and editable ultrasound image synthesis. Medical Image Analysis  \textbf{79},
   102461 (2022)

\bibitem{liu2022deep}
Liu, J., Zhou, X., Lin, H., Lu, X., Zheng, J., Xu, E., Jiang, D., Zhang, H.,
  Yang, X., Zhong, J., et~al.: Deep learning based on carotid transverse b-mode
  scan videos for the diagnosis of carotid plaque: a prospective multicenter
  study. European Radiology pp. 1--10 (2022)

\bibitem{liu2022video}
Liu, Z., Ning, J., Cao, Y., Wei, Y., Zhang, Z., Lin, S., Hu, H.: Video swin
  transformer. In: Proceedings of the IEEE/CVF conference on computer vision
  and pattern recognition. pp. 3202--3211 (2022)

\bibitem{peng2021conformer}
Peng, Z., Huang, W., Gu, S., Xie, L., Wang, Y., Jiao, J., Ye, Q.: Conformer:
  Local features coupling global representations for visual recognition. In:
  Proceedings of the IEEE/CVF international conference on computer vision. pp.
  367--376 (2021)

\bibitem{selvaraju2017grad}
Selvaraju, R.R., Cogswell, M., Das, A., Vedantam, R., Parikh, D., Batra, D.:
  Grad-cam: Visual explanations from deep networks via gradient-based
  localization. In: Proceedings of the IEEE international conference on
  computer vision. pp. 618--626 (2017)

\bibitem{wang2018non}
Wang, X., Girshick, R., Gupta, A., He, K.: Non-local neural networks. In:
  Proceedings of the IEEE conference on computer vision and pattern
  recognition. pp. 7794--7803 (2018)

\bibitem{xu2021vitae}
Xu, Y., Zhang, Q., Zhang, J., Tao, D.: Vitae: Vision transformer advanced by
  exploring intrinsic inductive bias. Advances in Neural Information Processing
  Systems  \textbf{34},  28522--28535 (2021)

\bibitem{yan2022multiview}
Yan, S., Xiong, X., Arnab, A., Lu, Z., Zhang, M., Sun, C., Schmid, C.:
  Multiview transformers for video recognition. In: Proceedings of the IEEE/CVF
  Conference on Computer Vision and Pattern Recognition. pp. 3333--3343 (2022)

\bibitem{yang2020temporal}
Yang, C., Xu, Y., Shi, J., Dai, B., Zhou, B.: Temporal pyramid network for
  action recognition. In: Proceedings of the IEEE/CVF conference on computer
  vision and pattern recognition. pp. 591--600 (2020)

\end{thebibliography}

\setcounter{table}{0} 
\renewcommand{\thetable}{S\Roman{table}}
\renewcommand\tablename{Supplementary Material: Table}
\begin{table}[h]
    \centering
    \belowrulesep=0pt
    \aboverulesep=0pt
    \caption{Architecture of \M, where each stage (\textit{c1-c5}) mainly includes a 3D Conv (mark in blue) and AMDF Trans (mark in red) blocks, except 3D Conv Sub-block in \textit{c1}. Residual connections are performed in all 3D Conv Sub-blocks. Kernels are denoted by $\left\{T \times S^2, C\right\}$ for temporal, spatial, and channel sizes. Strides denote $\left\{\text{temporal stride}, \text{spatial stride}^2\right\}$. MHSA-12, MHCA-12 denote the multi-head self-attention and cross attention with heads 12 in AMDF Trans blocks, respectively. Arrows in 3D FCU column show the flow of feature.}
    \begin{tabular}{c}
        \includegraphics[scale=1]{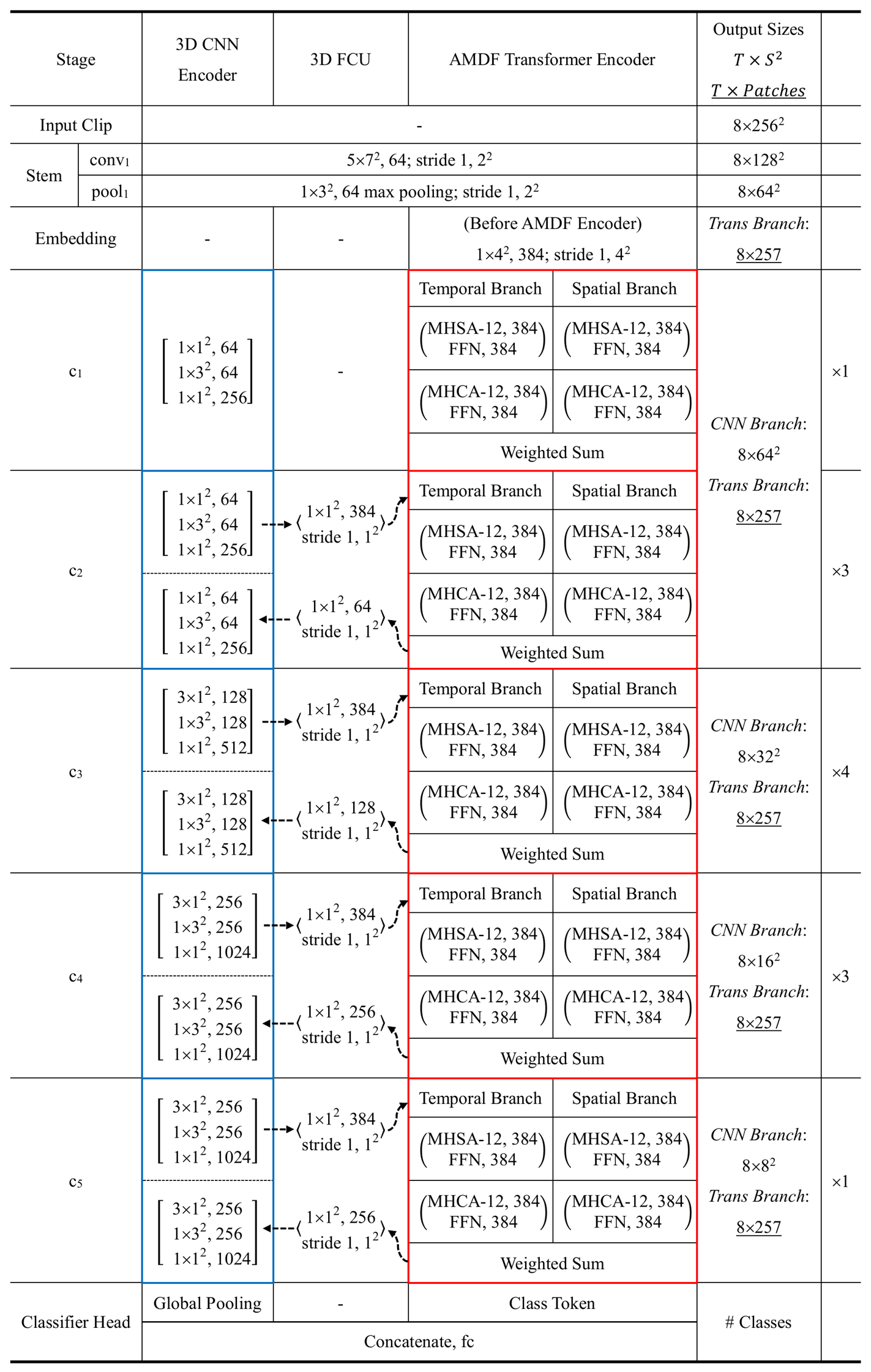}
    \end{tabular}
    \phantomsection
    \label{tab:supp}
\end{table}

\end{document}